\documentclass[a4paper,final]{article}
\usepackage[dvips]{graphicx}
\usepackage{amsmath} 
\usepackage{amssymb} 

\makeatletter

\renewenvironment{abstract}
  {\normalfont
    \list{}{\labelwidth0pt
      \leftmargin0pt \rightmargin\leftmargin
      \listparindent\parindent \itemindent0pt
      \parsep0pt
      
    }
    \item[\hskip\labelsep\bfseries\abstractname\enspace --] \itshape
}{
  \endlist}

\newcommand{\keywordsname}{Keywords}
\newenvironment{keywords}
  {\normalfont
    \list{}{\labelwidth0pt
      \leftmargin0pt \rightmargin\leftmargin
      \listparindent\parindent \itemindent0pt
      \parsep0pt
      }
    \item[\hskip\labelsep\bfseries\keywordsname:]}{\endlist}

\begin{document}

\pagestyle{myheadings}
\markboth{}{}

\title{Enrichment of Qualitative Beliefs for Reasoning under Uncertainty}

\author{
Xinde Li and Xinhan Huang\\
Int. Cont. and Robotics Lab., Dept. of Control Science and Eng.\\
Huazhong Univ of Sci. and Tech.\\
Wuhan 430074, China\\
xdl825@163.com\\
\and Florentin Smarandache\\
Department of Mathematics\\
University of New Mexico\\
Gallup, NM 87301, U.S.A.\\
smarand@unm.edu\\
\and Jean Dezert\\
ONERA\\
29 Av. de la  Division Leclerc \\
92320 Ch\^{a}tillon, France.\\
Jean.Dezert@onera.fr
}

\date{}

\maketitle

\begin{abstract}
This paper deals with enriched qualitative belief functions for reasoning under uncertainty and for combining information expressed in natural language through linguistic labels. In this work, two possible enrichments (quantitative and/or qualitative) of linguistic labels are considered and operators (addition, multiplication, division, etc) for dealing with them are proposed and explained. We  denote them $qe$-operators, $qe$ standing for ``qualitative-enriched" operators. These operators can be seen as a direct extension of the classical qualitative operators ($q$-operators) proposed recently in the Dezert-Smarandache Theory of plausible and paradoxist reasoning (DSmT). $q$-operators are also justified in details in this paper. The quantitative enrichment of linguistic label is a numerical supporting degree in $[0,\infty)$, while the qualitative enrichment takes its values in a finite ordered set of linguistic values. Quantitative enrichment is less precise than qualitative enrichment, but it is expected more close with what human experts can easily provide when expressing linguistic labels with supporting degrees. Two simple examples are given to show how the fusion of qualitative-enriched belief assignments can be done.
\end{abstract}

\begin{keywords}
Information fusion, Qualitative beliefs, DSmT, DST.
\end{keywords}

\section{Introduction}

Qualitative methods for reasoning under uncertainty have gained  more and more attention by Information Fusion community, especially by the researchers and system designers working in the development of modern multi-source systems for defense, robotics and so on. This is because traditional methods based only on quantitative representation and analysis are not able to completely satisfy adequately the need of the development of science and technology integrating at higher fusion levels human beliefs and reports in complex systems. Therefore qualitative knowledge representation becomes more and more important and necessary in next generations of (semi) intelligent automatic and autonomous systems.\\

For example, Wagner et al. \cite{Wagner} consider that although recent robots have powerful
sensors and actuators, their abilities to show intelligent behavior
is often limited because of lacking of appropriate spatial
representation. Ranganathan et al. \cite{Ranganathan} describe a
navigation system for a mobile robot which must execute motions in a
building, the environment is represented by a topological model
based on a Generalized Voronoi Graph (GVG) and by a set of visual
landmarks. A qualitative self-localization method for indoor
environment using a belt of ultrasonic sensors and a camera is
proposed. Moratz et al. \cite{Moratz} point out that
qualitative spatial reasoning (QSR) abstracts metrical details of
the physical world, of which two main directions are topological
reasoning about regions and reasoning about orientations of point
configurations. So, because concrete problems need a combination of
qualitative knowledge of orientation and qualitative knowledge of
distance, they present a calculus based on ternary relations where
they introduce a qualitative distance measurement based on two of
the three points. Duckham et al. \cite {Duckham} explore the
development and the use of a qualitative reasoning system based on
a description logic for providing the consistency between different
geographic data sets. Their research results suggest that further
work could significantly increase the level of automation for many
geographic data integration tasks.\\

Recently, Smarandache and Dezert in \cite{SDBook} (Chap. 10) give a
detailed introduction of major works for qualitative
reasoning under uncertainty. Among important works in this field, one must mention George Polya
 who first attempted in 1954 to find a formal characterization of qualitative human reasoning,
then followed by Lotfi Zadeh's works \cite{Zadeh1,Zadeh2}. Later, Wellman
\cite{Wellman} proposed a general characterization of qualitative
probability to relax precision in representation and reasoning
within the probabilistic framework, in order to develop Qualitative
Probabilistic Networks (QPN). Wong and Lingras \cite {Wong} have proposed a
method for generating basic belief functions from preference relations between each pair of propositions be specified
qualitatively based on Dempster-Shafer Theory (DST) \cite{Shafer_1976}. Parsons
\cite{Parsons1,Parsons2} then proposed a qualitative Dempster-Shafer
Theory, called Qualitative Evidence Theory (QET) by using techniques from qualitative reasoning. This approach seems however to have been abandoned by Parsons in favor of  qualitative probabilistic reasoning (QPR).  In 2004, Brewka et al. \cite {Brewka} have proposed a Qualitative Choice Logic (QCL), which is a propositional logic for representing alternative, ranked options for problem solutions. This logic adds to classical propositional logic a new connective called ordered disjunction, that is, if possible$ A$, but if $A$ is not possible then at least $B$. The semantics of qualitative choice logic is based on a preference relation among models. Very recently, Badaloni and Giacomin \cite{SM} integrate the ideas of flexibility and uncertainty into Allen's interval-based temporal framework and define a new formalism, called $IA^{fuz}$, which extends classical Interval Algebra (IA) to express qualitative fuzzy constraints between intervals.\\

In \cite{SDBook}, Smarandache and Dezert introduce a definition of qualitative basic belief assignment (qbba or just qm - standing for qualitative mass), and they propose an extension of quantitative fusion rules developed in DSmT framework for combining directly qbba's without mapping linguistic labels into numbers, and thus computing directly with words. Such extension (mainly the qualitative extension of DSmC, DSmH and PCR5 rules - see \cite{SDBook}) is based on the definition of new operators (addition, multiplication, etc) on linguistic labels which are called $q$-operators. In this work, we propose to enrich the original definition of qualitative basic belief assignment (qbba) into two possible different ways, quantitatively and qualitatively. These enrichments yields to the definition new linguistic operators for these new types of enriched qbba's. We will denote them $qe$-operators. \\

The first qbba enrichment consists in associating a quantitative (numerical) supporting degree in $[0,\infty)$ given a body of evidence/source to each linguistic label. Such enrichment allows to take into account and mix (when available) some numerical extra knowledge about the reliability/trustability of the linguistic label committed to propositions of the frame of discernment. The second possible enrichment is purely qualitative in order to fit more closely with what human experts are expected to provide in reality when enriching their linguistic labels using natural language.\\

This paper is organized as follows: In section \ref{sec2}, we remind briefly the basics of DSmT. In section \ref{sec3}  we present and justify in details the $q$-operators, in order to get ready for introducing new enriched qualitative-enriched (qe) operators in sections \ref{sec4}. In section \ref{sec6}, we illustrate through very simple examples how these operators can be used for combining enriched qualitative beliefs. Concluding remarks are then given in \ref{Conclusions}.

\section{Basics of DSmT for quantitative beliefs}
\label{sec2}

Let $\Theta=\{\theta_1,\theta_2,\cdots,\theta_n\}$ be a finite set of $n$ elements $\theta_i$, $i=1,\ldots,n$ assumed to be exhaustive. $\Theta$ corresponds to the frame of discernment of the problem under consideration. In general (unless introducing some integrity constraints), we assume that elements of $\Theta$  are non exclusive in order to deal with vague/fuzzy and relative concepts \cite{SDBook2}. This is the so-called {\it{free-DSm model}} which is denoted by  $\mathcal{M}^f(\Theta)$. In DSmT framework, there is no need to work on refined frame $\Theta_{ref}$ consisting in a (possibly finer) discrete finite set of exclusive and exhaustive hypotheses which is usually referred as {\it{Shafer's model}} $\mathcal{M}^0(\Theta)$ in literature, because DSm rules of combination work for any models of the frame, i.e. the  free DSm model, Shafer's model or any hybrid model. The hyper-power set (Dedekind's lattice) $D^\Theta$ is defined as the set of all compositions built from elements of $\Theta$ with $\cup$ and $\cap$ ($\Theta$ generates
 $D^\Theta$ under $\cup$ and $\cap$) operators such that
\begin{enumerate}
\item[a)]$\emptyset,\theta_1,\theta_2,\cdots,\theta_n\in D^\Theta$.
\item[b)]If $A,B\in D^\Theta$, then $A\cap B\in D^\Theta$ and
$A\cup B\in D^\Theta$.
\item[c)]No other elements belong to $D^\Theta$, except those obtained by using rules a) or b).
\end{enumerate}

A (quantitative) basic belief assignment (bba) expressing the belief committed to the elements of $D^\Theta$ by a given source/body of evidence $S$ is a mapping function $m(\cdot)$: $D^\Theta \rightarrow [0,1]$ such that:

\begin{equation}
m(\emptyset)=0 \qquad \mbox{and}\qquad \sum_{A\in D^\Theta} m(A)=1
\end{equation}

Elements $A\in D^\Theta$ having $m(A)>0$ are called {\it{focal elements}} of the bba $m(.)$. The general belief function and plausibility functions are defined respectively in almost the same
manner as within the DST \cite{Shafer_1976}, i.e.
\begin{equation}
Bel(A)=\sum\limits_{B\in D^\Theta, B\subseteq A} m(B)
\label{eq:Eq1}
\end{equation}
\begin{equation}
Pl(A)=\sum\limits_{B\in D^\Theta, B\cap A\neq \emptyset}m(B)
\label{eq:Eq2}
\end{equation}

The main concern in information fusion is the combination of sources of evidence and the efficient management of conflicting and uncertain information. DSmT offers several fusion rules, denoted by the generic symbol $\oplus$, for combining basic belief assignments. The simplest one, well adapted when working with the free-DSm\footnote{We call it {\it{free}} because no integrity constraint is introduced in such model.} model $\mathcal{M}^f(\Theta)$ and called DSmC (standing for {\it{DSm Classical rule}}) is nothing but the conjunctive fusion operator of bba's defined over the hyper-power set $D^\Theta$. Mathematically, DSmC for the fusion of $k\geq 2$ sources of evidence is defined by $m_{\mathcal{M}^f(\Theta)}(\emptyset)=0$ and $\forall A\neq\emptyset \in D^\Theta$,

\begin{align}
m_{\mathcal{M}^f(\Theta)}(A) & \triangleq [m_1\oplus\cdots\oplus m_k](A)\nonumber\\
m_{\mathcal{M}^f(\Theta)}(A) & =\sum\limits_{{X_1,\cdots,X_k\in D^\Theta}\atop {X_1
\cap\cdots\cap X_k = A}} \prod\limits_{s=1}^k m_s(X_s)
\label{eq:Eq3}
\end{align}

When working with hybrid models and/or Shafer's model $\mathcal{M}^0(\Theta)$, other rules for combination must be used for taking into account integrity constraints of the model (i.e. some exclusivity constraints and even sometimes no-existing constraints in dynamical problems of fusion where the model and the frame can change with time). For  managing efficiently the conflicts between sources of evidence, DSmT proposes mainly two alternatives to the classical Dempster's rule of combination \cite{Shafer_1976} for working efficiently with (possibly) high conflicting sources. The first rule proposed in \cite{SDBook2} was the DSm hybrid rule (DSmH) of combination which offers a prudent/pessimistic way of redistributing partial conflicting mass. The basic idea of DSmH is to redistribute the partial conflicting mass to corresponding partial ignorance. For example: let's consider only two sources with two bba's $m_1(.)$ and $m_2(.)$, if $A\cap B=\emptyset$ is an integrity constraint of  the model of $\Theta$ and if $m_1(A)m_2(B)>0$, then $m_1(A)m_2(B)$ will be transferred to $A\cup B$ through DSmH. The general formula for DSmH is quite complicated and can be found in \cite{SDBook2} and is not reported here due to space limitation. DSmH is actually a natural extension of Dubois \& Prade's rule of combination \cite{DP} which allows also to work with dynamical changes of the frame and its model. A much more precise fusion rule, called Proportional Conflict Redistribution rule no. 5 (PCR5) has been developed recently in \cite{SDBook} for transferring more efficiently  all partial conflicting masses. Basically, the idea of PCR5 is to transfer the conflicting mass only to the elements involved in the conflict and proportionally to their individual masses. For example: let's assume as before only two sources with bba's $m_1(.)$ and $m_2(.)$, $A\cap B=\emptyset$ for the model of $\Theta$ and $m_1(A)=0.6$ and $m_2(B)=0.3$. Then with PCR5, the partial conflicting mass $m_1(A)m_2(B)=0.6\cdot 0.3=0.18$ is redistributed to $A$ and $B$ only with the following proportions respectively: $x_A=0.12$ and $x_B=0.06$ because the proportionalization requires
$$\frac{x_A}{m_1(A)}=\frac{x_B}{m_2(B)}= \frac{m_1(A)m_2(B)}{m_1(A)+m_2(B)}=\frac{0.18}{0.9}=0.2$$
General PCR5 fusion formula for the combination of $k\geq 2$ sources of evidence can be found in \cite{SDBook}.

\section{Extension of DSmT for qualitative beliefs}
\label{sec3}

In order to compute with words (i.e. linguistic labels) and
qualitative belief assignments instead of quantitative belief
assignments\footnote{$G^\Theta$ is the generic notation for the
hyper-power set taking into account all integrity constraints (if
any) of the model. For example, if one considers a free-DSm model
for $\Theta$ then $G^\Theta=D^\Theta$. If Shafer's model is used
instead then $G^\Theta=2^\Theta$ (the classical power-set).} over
$G^\Theta$, Smarandache and Dezert have defined in \cite{SDBook} a
{\it{qualitative basic belief assignment}} $qm(.)$ as a mapping
function from $G^\Theta$ into a set of linguistic labels
$L=\{L_0,\tilde{L},L_{n+1}\}$ where $\tilde{L}=\{L_1,\cdots,L_n\}$
is a finite set of linguistic labels  and where $n \ge 2$ is an
integer. For example, $L_1$ can take the linguistic value ``poor",
$L_2$ the linguistic value ``good", etc. $\tilde{L}$ is endowed with
a total order relationship $\prec$, so that $L_1 \prec L_2 \prec
\cdots \prec L_n $. To work on a true closed linguistic set $L$
under linguistic addition and multiplication operators, Smarandache
and Dezert extended naturally $\tilde{L}$ with two extreme values
$L_0=L_{\min}$ and $ L_{n+1}=L_{\max}$, where $L_0$ corresponds to the minimal
qualitative value and $L_{n+1}$ corresponds to the maximal
qualitative value, in such a way that $L_0 \prec L_1 \prec L_2 \prec
\cdots \prec L_n  \prec L_{n+1} $, where $\prec$ means inferior to, or less (in quality) than, or smaller than, etc.  Labels $L_0$, $L_1$, $L_2$, \ldots, $L_n$, $L_{n+1}$ are said {\it{linguistically equidistant}} if:
$L_{i+1} - L_i = L_i - L_{i-1}$ for all $i = 1, 2, \ldots, n$ where the definition of subtraction of labels is given in  the sequel by \eqref{eq:qsub}. In the sequel $L_i \in L$ are assumed linguistically equidistant\footnote{If the labels are not equidistant, the q-operators still work, but they are less accurate.} labels such that we can make an isomorphism between $L = \{L_0, L_1, L_2, \ldots, L_n, L_{n+1}\}$ and $\{ 0, 1/(n+1), 2/(n+1), \ldots, n/(n+1), 1\}$, defined as $L_i = i/(n+1)$ for all $i = 0, 1, 2,\ldots, n, n+1$. Using this isomorphism, and making an analogy to the classical operations of real numbers, we are able to define the following qualitative operators (or $q$-operators for short):

\begin{itemize}
\item $q$-addition of linguistic labels
\begin{equation}
L_i+L_j=\frac{i}{n+1}+ \frac{j}{n+1}=\frac{i+j}{n+1}=L_{i+j}
\label{eq:q-addition}
\end{equation}
\noindent but of course we set the restriction that $i+j < n+1$; in the case when $i+j \geq n+1$ we restrict $L_{i+j} = L_{n+1}$. So this is the justification of the qualitative addition we have defined. 

\item $q$-multiplication of linguistic labels\footnote{The $q$-multiplication of two linguistic labels defined here can be extended directly to the multiplication of $n>2$ linguistic labels. For example the product of three linguistic label will be defined as $L_i \times L_j  \times L_k = L_{[(i\cdot j\cdot k)/(n+1)(n+1)]}$, etc.}

\begin{itemize}
\item[a)] Since $L_i \times L_j= \frac{i}{n+1}\cdot  \frac{j}{n+1} = \frac{(i\cdot j)/(n+1)}{n+1}$, the best approximation would be $L_{[(i\cdot j)/(n+1)]}$, where $[x]$ means the closest integer to $x$, i.e.
\begin{equation}
L_i \times L_j = L_{[(i\cdot j)/(n+1)]}
\label{eq:qmult}
\end{equation}
For example, if we have $L_0$, $L_1$, $L_2$, $L_3$, $L_4$, $L_5$, corresponding to respectively $0$, $0.2$, $0.4$, $0.6$, $0.8$, $1$, then $L_2 \cdot L_3 = L_{[(2\cdot 3)/5]} = L_{[6/5]} = L_{[1.2] }= L_1$; using numbers: $0.4\cdot 0.6 = 0.24 \approx 0.2 = L_1$; also $L_3 \cdot L_3 = L_{[(3\cdot 3)/5]} = L_{[9/5]} = L_{[1.8]} = L_2$; using numbers $0.6 \cdot 0.6 = 0.36 \approx 0.4 = L_2$.
\item[b)] A simpler approximation of the multiplication, but less
accurate (as proposed in \cite{SDBook}) is thus
\begin{equation}
L_i \times L_j = L_{\min\{i,j\}} \label{eq:qmultsimpler}
\end{equation}
\end{itemize}

\item Scalar multiplication of a linguistic label

Let $a$ be a real number. We define the multiplication of a linguistic label by a scalar as follows:

\begin{equation}
a \cdot L_i =\frac{a\cdot i}{n+1} \approx
\begin{cases}
L_{[a\cdot i]} & \text{if} \ [a\cdot i]\geq  0,\\
L_{-[a\cdot i]} & \text{otherwise}.
\end{cases}
\label{eq:sqmult}
\end{equation}

\item Division of linguistic labels

\begin{itemize}
\item[a)]  Division as an internal operator:  $/ : L \times  L \rightarrow  L$.
Let $j\neq 0$, then
\begin{equation}
L_i / L_j  =
\begin{cases}
L_{[(i/j)  \cdot (n+1)]} & \text{if} [(i/j) \cdot (n+1)] < n+1,\\
L_{n+1} & \text{otherwise}.
\end{cases}
\label{eq:sqdiv}
\end{equation}

The first equality in \eqref{eq:sqdiv} is well justified because when $ [(i/j) \cdot (n+1)] < n+1$, one has $$L_i / L_j  = \frac{i/(n+1)}{j/(n+1)}=\frac{(i/j)\cdot (n+1)}{n+1}=L_{[(i/j)  \cdot (n+1)]}$$
For example, if we have $L_0$, $L_1$, $L_2$, $L_3$, $L_4$, $L_5$, corresponding to respectively $0$, $0.2$, $0.4$, $0.6$, $0.8$, $1$, then: $L_1 / L_3 = L_{[(1/3)\cdot  5]}=L_{[5/3]}=L_{[1.66]} \approx  L_2$. $L_4 / L_2 = L_{[(4/2)\cdot 5]}=L_{[2\cdot 5]}=L_{\max}=L_5$ since $10>5$.

\item[b)] Division as an external operator: $\oslash : L \times  L \rightarrow  \mathbb{R}^+$.
Let $j\neq 0$. Since $L_i \oslash L_j = (i/(n+1)) / (j/(n+1)) = i/j$, we simply define 
\begin{equation}
L_i \oslash L_j  = i/j
\label{eq:sqextdiv}
\end{equation}
\noindent
Justification of b): when we divide say $L_4 / L_1$ in the above example, we get $0.8/0.2 = 4$, but no label is corresponding to number 4 which is not even in the interval $[0,1]$, hence in the division as an internal operator we need to get as response a label, so in our example we approximate it to $L_{\max}=L_5$, which is a very rough approximation!  So, depending on the fusion combination rules, it might better to consider the qualitative division as an external operator, which gives us the exact result.
\end{itemize}

\item $q$-subtraction of linguistic labels:  $- : L \times  L \rightarrow \{L, -L\}$, 

\begin{equation}
 L_i - L_j =
\begin{cases}
L_{i-j} & \text{if} \quad i \geq  j,\\
- L_{j-i} &  \text{if} \quad i <  j.
\end{cases}
\label{eq:qsub}
\end{equation}
\noindent
where $ -L =  \{ -L_1, -L_2, \ldots , -L_n, -L_{n+1} \}$. 
The $q$-subtraction above is well justified since when $i \geq  j$, one has $L_i - L_j =  \frac{i}{n+1} - \frac{j}{n+1} =  \frac{i-j}{n+1}$.

\end{itemize}

The above qualitative operators are logical, justified due to the isomorphism between the set of linguistic equidistant labels and a set of equidistant numbers in the interval $[0,1]$.  These qualitative operators are built exactly on the track of their corresponding numerical operators, so they are more mathematical than the ad-hoc definition of qualitative operators proposed so far in the literature. They are similar to the PCR5 combination numerical rule with respect to other fusion combination numerical rules based on the conjunctive rule. But moving to the enriched label qualitative operators the accuracy decreases.\\

\noindent
{\bf{Remark about doing multi-operations on labels}}: When working with labels, no matter how many
operations we have, the best (most accurate) result is obtained if we do only one approximation, and that one should be just at the very end. For example, if we have to compute terms like $L_iL_jL_k / (L_p+L_q)$ as for qPCR5 (see example in section \ref{sec6}), we compute all operations as defined above, but without any approximations (i.e. not even calculating the integer part of indexes, neither replacing by $n+1$ if the intermediate results is bigger than $n+1$), so:

\begin{equation}
\frac{L_iL_jL_k }{L_p+L_q}=\frac{L_{(ijk)/(n+1)^2}}{L_{p+q} }= L_{{ \frac{(ijk) / {(n+1)^2}}{p+q}\cdot (n+1)}} = L_{ \frac{(ijk)/(n+1)}{p+q} } 
=L_{ \frac{ijk}{(n+1)(p+q)} }
\label{eq12}
\end{equation}

\noindent and now, when all work is done, we compute
the integer part of the index, i.e. $[\frac{ijk}{(n+1)(p+q)}]$ or replace it by $n+1$ if the final result is bigger than $n+1$.
Therefore, the term $L_iL_jL_k / (L_p+L_q)$ will take the linguistic value $L_{n+1}$ whenever $[\frac{ijk}{(n+1)(p+q)}] > n+1$. This method also insures us of a unique result, and it is mathematically closer to the result that would be
obtained if working with corresponding numerical masses. Otherwise, if one does approximations either at the beginning or after each operation or in the middle of calculations, the inaccuracy propagates (becomes bigger and bigger) and we obtain different results, depending on the places where the approximations were done.

\section{Quasi-normalization of $qm(.)$}

There is no way to define a normalized $qm(.)$, but a qualitative quasi-normalization \cite{SDBook,QBCR} is nevertheless possible when considering equidistant linguistic labels because in such case, $qm(X_i) = L_i$, is equivalent to a quantitative mass $m(X_i) = i/(n+1)$ which is normalized if 
$$\sum_{X\in D^\Theta} m(X)= \sum_{k} i_k/(n+1)=1$$
\noindent
but this one is equivalent to 
$$\sum_{X\in D^\Theta} qm(X)= \sum_{k} L_{i_k}=L_{n+1}$$
\noindent
In this case, we have a {\it{qualitative normalization}}, similar to the (classical) numerical normalization. But, if the previous labels $L_0$, $L_1$, $L_2$, $\ldots$, $L_n$, $L_{n+1}$ from the set $L$ are not equidistant, so the interval $[0, 1]$ cannot be split into equal parts according to the distribution of the labels, then it makes sense to consider a {\it{qualitative quasi-normalization}}, i.e. an approximation of the (classical) numerical normalization for the qualitative masses in the same way:
 $$\sum_{X\in D^\Theta} qm(X)=L_{n+1}$$
\noindent
In general, if we don't know if the labels are equidistant or not, we say that a qualitative mass is quasi-normalized when the above summation holds. In the sequel, for simplicity, one assumes to work with quasi-normalized qualitative basic belief assignments.\\

From these very simple qualitative operators, it is thus possible to extend directly the DSmH fusion rule for combining qualitative basic belief assignments by replacing classical addition and multiplication operators on numbers with those for linguistic labels in DSmH formula. In a similar way, it is also possible to extend PCR5 formula as shown with detailed examples in \cite{SDBook} and in section \ref{sec6} of this paper. In the next section, we propose new qualitative-enriched (qe) operators for dealing with enriched  linguistic labels which mix the linguistic value with either quantitative/numerical supporting degree or qualitative supporting degree as well. The direct qualitative discounting (or reinforcement) is motivated by the fact that in general human experts provide more easily qualitative values than quantitative values when analyzing complex situations. \\

In this paper, both {\it{quantitative enrichments}} and {\it{qualitative enrichments}} of linguistic labels are considered and unified through same general $qe$-operators. The quantitative enrichment is based directly on the percentage of discounting (or reinforcement) of any linguistic label. This is what we call a Type 1 of enriched labels. 
The qualitative enrichment comes from the idea of direct qualitative discounting (or reinforcement) and constitutes the Type 2 of enriched labels. 

\section{$qe$-operators}
\label{sec4}

We propose to improve the previous $q$-operators for dealing now
with enriched qualitative beliefs provided from human experts.
We call these operators the $qe$-operators. The basic idea is
to use ``enriched"-linguistic labels denoted $L_i(\epsilon_i)$,
where $\epsilon_i$ can be either a numerical supporting degree in $[0,\infty)$ or a qualitative supporting degre taken its value in a given (ordered) set $X$ of linguistic labels. $L_i(\epsilon_i)$ is called the qualitative enrichment\footnote{Linguistic labels without enrichment (discounting
or reinforcement) as those involved in $q$-operators are said {\it
{classical}} or being of Type 0.} of $L_i$. When $\epsilon_i \in [0,\infty)$, $L_i(\epsilon_i)$ is called an enriched label of Type 1, whereas when $\epsilon_i \in X$, $L_i(\epsilon_i)$ is called an enriched label of Type 2. The (quantitative or qualitative) quantity $\epsilon_i$ characterizes the weight of
reinforcing  or discounting expressed by the source when using label
$L_i$ for committing its qualitative belief to a given proposition
$A\in G^\Theta$. It can be interpreted as a second order type of
linguistic label which includes both the linguistic value itself but
also the associated degree of confidence expressed by the source. The values of
$\epsilon_i$ express the expert's attitude (reinforcement, neutral,
or discounting) to a certain proposition when using a given linguistic label for expressing its qualitative belief assignment.\\

For example with enriched labels of Type 1, if the label $L_1\triangleq L_1(1)$ represents the
linguistic variable \textit{Good}, then $L_1(\epsilon_1)$ represents
either the reinforced or discounted $L_1$ value which depends on the
value taken by $\epsilon_1$. In this example, $\epsilon_1$ represents
the (numerical) supporting degree of  the linguistic value $L_1=\text{Good}$. If  $\epsilon_1 = 1.2$, then we say that the linguistic value $L_1=\text{Good}$ has been reinforced by $20\%$ with respect to its nominal/neutral supporting degree. If $\epsilon_1 =0.4$, then it means that the linguistic value $L_1$ is discounted 60\% by the source. \\

With enriched labels of Type 2, if one chooses by example $X=\{NB,NM,NS,O,PS,PM,PB\}$, where elements of $X$ have the following meaning:  $NB\triangleq \text{``negative big''}$, $NM\triangleq \text{``negative medium''}$, $NS\triangleq$``negative small'', $O\triangleq \text{``neutral''}$ (i.e. no discounting, neither reinforcement), $PS\triangleq \text{``positive small''}$, $PM\triangleq$``positive medium'' and $PB\triangleq \text{``positive big''}$. Then, if the label $L_1\triangleq L_1(O)$ represents the linguistic variable \textit{Good}, then $L_1(\epsilon_1)$, $\epsilon_1 \in X$,  represents either the qualitative reinforced or discounted $L_1$ value which depends on the
value taken by $\epsilon_1$ in $X$. $\epsilon_1=O$ means a neutral qualitative supporting degree corresponding to $\epsilon_1=1$ for enriched label of Type 1. $\epsilon_1$ represents
the qualitative supporting degree of  the linguistic value $L_1=\text{Good}$. If  $\epsilon_1 = PS$, then we say that the linguistic value $L_1=\text{Good}$ has been reinforced a little bit positively with respect to its nominal/neutral supporting degree. If $\epsilon_1 =NB$, then it means that the linguistic value $L_1$ is discounted slightly and negatively by the source.\\

We denote by $\tilde{L}(\epsilon)$ any given set of (classical/pure)
linguistic labels $\tilde{L}=\{L_1, L_2,\ldots, L_n\}$ endowed with
the supporting degree property (i.e. discounting, neutral and/or
reinforcement). In other words,
$$\tilde{L}(\epsilon)=\{L_1(\epsilon_1),
L_2(\epsilon_2),\ldots, L_n(\epsilon_n)\}$$
\noindent represents a given set of enriched linguistic
labels\footnote{In this formal notation, the quantities $\epsilon_1$, $\ldots$, $\epsilon_n$
represent any values in $[0,\infty)$ if the enrichment is quantitative (Type 1), or values in $X$ is we consider an qualitative enrichment (Type 2).}. We assume the same order relationship $\prec$ on
 $\tilde{L}(\epsilon)$ as the one defined on $\tilde{L}$. Moreover we extend $\tilde{L}(\epsilon)$
 with two extreme (minimal and maximal) enriched qualitative values $L_0(\epsilon)$ and $L_{n+1}(\epsilon)$ in order to get
 closed under $qe$-operators on $L(\epsilon)\triangleq\{L_0(\epsilon), \tilde{L}(\epsilon),L_{n+1}(\epsilon)\}$. For working with enriched labels (and then with qualitative enriched basic belief assignments), it is necessary to extend the previous $q$-operators in a consistent way. This is the purpose of our new $qe$-operators. \\
 
An enriched label $L_i(\epsilon_i)$ means that the source has discounted (or reinforced) the label $L_i$ by a quantitative or qualitative factor $\epsilon_i$. Similarly for $L_j(\epsilon_j)$. So we use the $q$-operators for  $L_i$, $L_j$ labels, but
for confidences we propose three possible versions:
If the confidence in $L_i$ is $\epsilon_i$ and the confidence in
$L_j$ is $\epsilon_j$, then the confidence in combining $L_i$ with
$L_j$ can be:\\

\begin{itemize}
\item[a)] either the average, i.e. $(\epsilon_i+\epsilon_j)/2$;
\item[b)] or $\min\{\epsilon_i, \epsilon_j\}$;
\item[c)] or we may consider a confidence interval as in statistics, so we get $[ \epsilon_{\min}, \epsilon_{\max} ] $, 
where $\epsilon_{\min} \triangleq \min \{\epsilon_i, \epsilon_j\}$ and $\epsilon_{\max}\triangleq \max \{\epsilon_i, \epsilon_j\}$; if $\epsilon_i=\epsilon_j$ then the confidence interval is reduced to a single point, $\epsilon_i$.\\
\end{itemize}

In the sequel, we denote by ``$c$" any of the above resulting confidence of combined enriched labels. All these versions coincide when $\epsilon_i=\epsilon_j=1$ (for Type 1) or when $\epsilon_i=\epsilon_j=O$ (for Type 2), i.e. where there is no reinforcement or no discounting of the linguistic label. However the confidence degree average operator (case a) ) is not associative, so in many cases it's inconvenient to use it. The best among these three, more prudent and easier to use, is the $\min$ operator. The confidence interval operator provides both a lower and a upper confidence level, so in an optimistic way, we may take at the end the midpoint of this confidence interval as a confidence level.\\

The new extended operators allowing working with enriched labels of Type 1 or Type 2 are then defined by:\\

\begin{itemize}
\item $qe$-addition of enriched labels

\begin{equation}
 L_i (\epsilon_i)+ L_j (\epsilon_j)=
\begin{cases}
L_{n+1}(c) \quad \text{if} \quad i +j\geq  n+1,\\
L_{i+j}(c) \quad \text{otherwise.}
\end{cases}
\label{eq:qpadd}
\end{equation}

\item $qe$-multiplication of linguistic labels

\begin{itemize}
\item[a)] As direct extension of \eqref{eq:qmult}, the multiplication of enriched labels is defined by
\begin{equation}
L_i(\epsilon_i)\times L_j(\epsilon_j)= L_{[(i\cdot j)/(n+1)]}(c)
\label{eq:qpmulta}
\end{equation}
\item[b)] as another multiplication of labels, easier, but less exact:
\begin{equation}
 L_i (\epsilon_i)\times L_j (\epsilon_j)= L_{\min\{i,j\}}(c)
\label{eq:qpmult}
\end{equation}
\end{itemize}

\item Scalar multiplication of a enriched label

Let $a$ be a real number. We define the multiplication of an enriched linguistic label by a scalar as follows:

\begin{equation}
a \cdot L_i (\epsilon_i)\approx
\begin{cases}
L_{[a\cdot i]}(\epsilon_i) & \text{if} \ [a\cdot i]\geq  0,\\
L_{-[a\cdot i]}(\epsilon_i) & \text{otherwise}.
\end{cases}
\label{eq:sqmultenriched}
\end{equation}

\item $qe$-division of enriched labels

\begin{itemize}
\item[a)]  Division as an internal operator:  

Let $j\neq 0$, then
\begin{equation}
\frac{L_i(\epsilon_i)}{L_j(\epsilon_j)}  =
\begin{cases}
L_{n+1}(c) \quad \text{if} \ [(i/j) \cdot (n+1)] \geq n+1,\\
L_{[(i/j)  \cdot (n+1)]} (c)\quad  \text{otherwise}.
\end{cases}
\label{eq:sqpdiv}
\end{equation}

\item[b)] Division as an external operator: 

Let $j\neq 0$, then we can also consider the division of enriched labels as external operator too as follows:

\begin{equation}
{L_i(\epsilon_i)}\oslash {L_j(\epsilon_j)}  = (i/j)_{\text{supp}(c)}
\label{eq:sqpextdiv}
\end{equation}
\noindent
The notation $(i/j)_{\text{supp}(c)}$ means that the numerical value $(i/j)$ is supported with the degree $c$.\\

\end{itemize}

\item $qe$-subtraction of enriched labels 

\begin{equation}
 L_i(\epsilon_i) - L_j(\epsilon_j) =
\begin{cases}
L_{i-j}(c)& \text{if} \quad i \geq  j,\\
- L_{j-i}(c) &  \text{if} \quad i <  j.
\end{cases}
\label{eq:qpsub}
\end{equation}
\noindent

\end{itemize}

These $qe$-operators with numerical confidence degrees are consistent with the classical qualitative
operators when $e_i=e_j =1$ since $c=1$ and $L_i(1)=L_i$ for
all $i$, and the $qe$-operators with qualitative confidence degrees are also consistent with the
classical qualitative operators when $e_i=e_j = O$ (this is letter ``O", not zero, hence the neutral qualitative confidence degree) since $c =O$ (neutral).

\section{Examples of qPCR5 fusion of qualitative belief assignments}
\label{sec6}

\subsection{Qualitative masses using quantitative enriched labels}

Let's consider a simple frame $\Theta=\{A,B\}$ with Shafer's model (i.e. $A\cap B=\emptyset$), two qualitative belief assignments $qm_1(\cdot)$ and $qm_2(\cdot)$, the set of ordered linguistic labels $L=\{L_0,L_1,L_2,L_3,L_4,L_5,L_6\}$, $n=5$, enriched with quantitative support degree (i.e. enriched labels of Type 1). For this example the (prudent) $\min$ operator for combining confidences proposed in section \ref{sec4} (case b) ) is used, but other methods a) and c) can also be applied.We consider the following qbba summarized in the Table \ref{Table1}:
\begin{table}[htbp]
\begin{center}
\begin{tabular}{|c|ccc|c|} \hline
   & $A$ & $B$ & $A\cup B$ & $A\cap B$\\ \hline   
 $qm_1(\cdot)$ & $L_1(0.3)$ & $L_2(1.1)$ & $L_3(0.8)$ & \\
  $qm_2(\cdot)$ & $L_4(0.6)$ & $L_2(0.7)$ & $L_0(1)$ & \\ \hline
  $qm_{12}(\cdot)$ & $L_3(0.3)$ & $L_2(0.7)$ & $L_0(0.8)$ & $L_1(0.3)$\\ \hline
 \end{tabular}
\end{center}
\caption{ $qm_1(\cdot)$, $qm_2(\cdot)$ and $qm_{12}(\cdot)$ with quantitative enriched labels}
\label{Table1}
\end{table}

\noindent
Note that $qm_1(\cdot)$ and $qm_2(\cdot)$ are quasi-normalized since $L_1+L_2+L_3=L_4+L_2+L_0=L_6=L_{\max}$.
The last raw of Table \ref{Table1}, corresponds to the result $qm_{12}(\cdot )$ obtained when applying the qualitative conjunction rule. The values for $qm_{12}(\cdot )$ are obtained using intermediate approximations as follows:

\begin{align*}
qm_{12}(A) &= qm_1(A)qm_2(A)+qm_1(A)qm_2(A\cup B) + qm_2(A)qm_1(A\cup B)\\
& = L_1(0.3)L_4(0.6)+L_1(0.3)L_0(1) +L_4(0.6)L_3(0.8)\\
& \approx L_{[(1\cdot 4)/6]}(\min\{0.3,0.6\}) + L_{[(0\cdot 1)/6]}(\min\{0.3,1\}) + L_{[(4\cdot 3)/6]}(\min\{0.6,0.8\})\\
& = L_{1}(0.3) + L_{0}(0.3)+ L_{2}(0.6) = L_{1+0+2}(\min\{0.3,0.3,0.6\}) = L_{3}(0.3)
\end{align*}
\begin{align*}
qm_{12}(B) &= qm_1(B)qm_2(B)+qm_1(B)qm_2(A\cup B)  + qm_2(B)qm_1(A\cup B)\\
& = L_2(1.1)L_2(0.7)+L_2(1.1)L_0(1) +L_2(0.7)L_3(0.8)\\
& \approx L_{[(2\cdot 2)/6]}(\min\{1.1,0.7\}) + L_{[(2\cdot 0)/6]}(\min\{1.1,1\}) + L_{[(2\cdot 3)/6]}(\min\{0.7,0.8\})\\
& = L_{1}(0.7) + L_{0}(1)+ L_{1}(0.7) = L_{1+0+1}(\min\{0.7,1,0.7\})= L_{2}(0.7)
\end{align*}

\begin{align*}
qm_{12}(A\cup B) &= qm_1(A\cup B)qm_2(A\cup B) = L_3(0.8)L_0(1)\\
& \approx L_{[(3\cdot 0)/6]}(\min\{0.8,1\})= L_{0}(0.8)
\end{align*}
\noindent
and the conflicting qualitative mass by
\begin{align*}
qm_{12}(\emptyset) &= qm_{12}(A\cap B) = qm_1(A)qm_2(B)+qm_2(A)qm_1(B) \\
& = L_1(0.3)L_2(0.7) + L_4(0.6)L_2(1.1)\\
& \approx L_{[(1\cdot 2)/6]}(\min\{0.3,0.7\})  + L_{[(4\cdot 2)/6]}(\min\{0.6,1.1\})\\
& = L_{0}(0.3) + L_{1}(0.6) = L_{0+1}(\min\{0.3,0.6\}) = L_{1}(0.3)
\end{align*}

\noindent
The resulted qualitative mass, $qm_{12}(\cdot )$, is (using intermediate approximations) quasi-normalized since $L_3+L_2+L_0+L_1=L_6=L_{\max}$.\\

\noindent
Note that, when the derivation of $qm_{12}(.)$ is carried out with the approximations done at the end (i.e. the best way to carry derivations), one gets for $qm_{12}(A)$ in a similar way as in \eqref{eq12}:

\begin{align*}
qm_{12}(A) &= qm_1(A)qm_2(A)+qm_1(A)qm_2(A\cup B) + qm_2(A)qm_1(A\cup B)\\
& = L_1(0.3)L_4(0.6)+L_1(0.3)L_0(1) +L_4(0.6)L_3(0.8)\\
&=(L_{1\times 4/6} + L_{1\times 0/6} + L_{4\times 3/6}) (\min\{0.3, 0.6, 0.3, 1, 0.6, 0.8\})\\
& = L_{4/6 + 0/6 + 12/6}(0.3) = L_{16/6}(0.3)  \approx L_{[16/6]}(0.3) = L_3(0.3)
\end{align*}

\noindent
Similarly:

\begin{align*}
qm_{12}(B) &= qm_1(B)qm_2(B)+qm_1(B)qm_2(A\cup B)  + qm_2(B)qm_1(A\cup B)\\
& = L_2(1.1)L_2(0.7)+L_2(1.1)L_0(1) +L_2(0.7)L_3(0.8)\\
&=L_{2\times 2/6 + 2\times 0/6 + 2\times 3/6}(\min\{1.1, 0.7, 1.1, 1, 0.7, 0.8\})\\
&=L_{4/6 + 0/6 + 6/6}(0.7) = L_{10/6}(0.7) \approx L_{[10/6]}(0.7) = L_2(0.7)
\end{align*}

\noindent
But for $qm_{12}(\emptyset)$, computed in a similar way as we did in \eqref{eq12}, one gets:

\begin{align*}
qm_{12}(\emptyset) &= qm_{12}(A\cap B) = qm_1(A)qm_2(B)+qm_2(A)qm_1(B) \\
&= L_1(0.3)L_2(0.7) + L_4(0.6)L_2(1.1)\\
&= L_{1\times 2/6 + 4\times 2/6}(\min\{0.3, 0.7, 0.6, 1.1\})\\
&=L_{10/6}(0.3) \approx L_{[10/6]}(0.3) = L_2(0.3)
\end{align*}

\noindent
Which is different from the previous case when we approximate each time and not only at the end.
So $qm_{12}(.)$ is not quasi-normalized in this way of calculation.\\

According to qPCR5 (see \cite{SDBook}), we need to redistribute the conflicting mass $L_1(0.3)$ to the elements involved in the conflict, $A$ and $B$, thus:
\begin{itemize}
\item[a)] $qm_1(A)qm_2(B)=L_1(0.3)L_2(0.7)=L_0(0.3)$ is redistributed back to $A$ and $B$ proportionally with respect to their corresponding qualitative masses put in this partial conflict, i.e. proportionally with respect to $L_1(0.3)$ and $L_2(0.7)$. But, since $L_0(0.3)$ is the null qualitative label (equivalent to zero for numerical masses), both $A$ and $B$ get $L_0$ with the minimum confidence, i.e. $L_0(0.3)$.
\item[b)] $qm_2(A)qm_1(B)=L_4(0.6)L_2(1.1)=L_1(0.6)$ is redistributed back to $A$ and $B$ proportionally with respect to their corresponding qualitative masses put in this partial conflict, i.e. proportionally with respect to $L_4(0.6)$ and $L_2(1.1)$, i.e.
$$
\frac{x_A}{L_4(0.6)}=\frac{y_B}{L_2(1.1)}=\frac{L_4(0.6)L_2(1.1)}{L_4(0.6) + L_2(1.1)}
$$
\noindent
whence using \eqref{eq12}, one gets
$$
x_A  =\frac{L_4(0.6)L_4(0.6)L_2(1.1)}{L_4(0.6) + L_2(1.1)}  = L_{[ ((4\cdot 4 \cdot 2)/6^2)/(4+2))(6)]}(\min\{0.6,0.6,1.1,0.6,1.1\})= L_{[ 8/9]}(0.6)=L_1(0.6)
$$
$$
y_B=\frac{L_2(1.1)L_4(0.6)L_2(1.1)}{L_4(0.6) + L_2(1.1)}  = L_{[ ((2\cdot 4 \cdot 2)/6^2)/(4+2))(6)]}(\min\{1.1,0.6,1.1,0.6,1.1\}) =L_{[ 4/9]}(0.6)=L_0(0.6)
$$
\noindent
Note that in this particular example, we get the same result if one uses the intermediate approximation as published in \cite{Li2007}, i.e.
$$
x_A  =L_4(0.6)\cdot \frac{L_4(0.6)L_2(1.1)}{L_4(0.6) + L_2(1.1)} \approx L_4(0.6)\cdot \frac{L_1(0.6)}{L_6(0.6)}  =  L_4(0.6)\cdot L_1(0.6)=L_{[(4\cdot 1)/6]}(\min\{0.6,0.6\})=L_1(0.6)
$$
$$
y_B=L_2(1.1)\cdot   \frac{L_4(0.6)L_2(1.1)}{L_4(0.6) + L_2(1.1)} \approx L_2(1.1)\cdot \frac{L_1(0.6)}{L_6(0.6)} = L_2(1.1)\cdot L_1(0.6)  = L_{[(2\cdot 1)/6]}(\min\{1.1,0.6\}) =L_0(0.6)
$$

\end{itemize}
\noindent
Thus, the result of the qPCR5 fusion of $qm_1(\cdot )$ with $qm_2(\cdot )$ is given by
$$
qm_{PCR5}(A) =L_3(0.3) + L_0(0.3) + x_A =L_3(0.3) + L_0(0.3) + L_1(0.6) = L_{3+0+1}(\min\{0.3,0.3,0.6\}) = L_4(0.3)
$$
$$
qm_{PCR5}(B)=L_2(0.7) + L_0(0.3) + y_B =L_2(0.7) + L_0(0.3)  + L_0(0.6) = L_{2+0+0}(\min\{0.7,0.3,0.6\}) = L_2(0.3)
$$
$$qm_{PCR5}(A\cup B) =L_0(0.8)$$
$$qm_{PCR5}(A\cap B) =L_0=L_0(1)$$

\noindent
This qualitative PCR5-combined resulting mass is also quasi-normalized\footnote{The confidence level/degree in the labels does not matter in the definition of quasi-normalization.} since $L_4+L_2+L_0+L_0=L_6=L_{\max}$.\\

\subsection{Qualitative masses with qualitative enriched labels}

Using qualitative supporting degrees (i.e. enriched labels of Type 2) taking their values in the linguistic set $X=\{NB,NM,NS,O,PS,PM,PB\}$, with $NB\prec NM\prec NS \prec O \prec PS \prec PM \prec PB$ we get similar result for this example. So, let's consider a frame $\Theta=\{A,B\}$ with Shafer's model and $qm_1(\cdot)$ and $qm_2(\cdot)$ chosen as in Table \ref{Table2}

\begin{table}[htbp]
\begin{center}
\begin{tabular}{|c|ccc|} \hline
   & $A$ & $B$ & $A\cup B$ \\ \hline   
 $qm_1(\cdot)$ & $L_1(NB)$ & $L_2(PS)$ & $L_3(NS)$ \\
  $qm_2(\cdot)$ & $L_4(NM)$ & $L_2(NS)$ & $L_0(O)$ \\ \hline
  \end{tabular}
\end{center}
\caption{ $qm_1(\cdot)$, $qm_2(\cdot)$ with qualitative enriched labels}
\label{Table2}
\end{table}

\noindent
The qualitative conjunctive  and PCR5 fusion rules are obtained with derivations identical to the previous ones, since $NB\prec NM\prec NS \prec O \prec PS \prec PM \prec PB$ and we associated $NB=0.3$ or less, $NM=[0.5,0.6]$, $NS=[0.7,0.8]$, $O=1$ and $PS=1.1$. The minimum operator on $X$(qualitative degrees) works similarly as on $\mathbb{R}^+$ (quantitative degrees). Thus, finally one gets results according to Table \ref{Table3}.

\begin{table}[htbp]
\begin{center}
\begin{tabular}{|c|ccc|c|} \hline
   & $A$ & $B$ & $A\cup B$ & $A\cap B$\\ \hline   
  $qm_{12}(\cdot)$ & $L_3(NB)$ & $L_2(NS)$ & $L_0(NS)$ & $L_1(NB)$\\ \hline
  $qm_{PCR5}(\cdot)$ & $L_4(NB)$ & $L_2(NB)$ & $L_0(NS)$ & $L_0(O)$\\ \hline
 \end{tabular}
\end{center}
\caption{Result obtained with qualitative conjunctive and PCR5 fuion rules}
\label{Table3}
\end{table}

\section{Conclusion}
\label{Conclusions}

With the recent development of qualitative methods for reasoning under uncertainty developed in Artificial Intelligence,
more and more experts and scholars have great interest on qualitative information fusion, especially those working in the
development of modern multi-source systems for defense, robot navigation, mapping, localization and path planning and so on. In this paper, we have proposed two possible enrichments (quantitative and/or qualitative) of linguistic labels and a simple and direct extension of the $q$-operators developed in the DSmT framework. We have also shown how to fuse qualitative-enriched belief assignments which can be expressed in natural language by human experts. Two illustrating examples have been presented in details to  explain how our qualitative-enriched operators ($qe$-operators) and qualitative PCR5 rule of combination work. Some research  in robotics of the application of $qe$-operators (with quantitative or qualitative supporting degrees) is under progress and will be presented in 	a forthcoming publication.

\section*{Acknowledgment}
This work is partially supported by the National Nature Science
Foundation of China (No.60675028).


\begin{thebibliography}{99}

\bibitem{SM}
S.~Badaloni and M.~Giacomin, ``The algebra $IA^{fuz}$: a framework
for qualitative fuzzy temporal reasoning,'' \emph{Artificial
Intelligence}, Vol. 170, No. 10, pp. 872--908, July 2006.

\bibitem{Brewka}
G.~Brewka, S.~Benferhat and D. L.~Berre, ``Qualitative choice
logic,'' \emph{Artificial Intelligence}, Vol.157, No. 1-2, pp.
203--237, August 2004.

\bibitem{DP}
D.~Dubois and H.~Prade, ``Representation and combination of uncertainty with belief functions and possibility measures,'' \emph{Computational Intelligence}, Vol. 4, pp. 244--264, 1988.

\bibitem{Duckham}
M.~Duckham, J.~Lingham, K.~Mason and M.~Worboys, ``Qualitative
reasoning about consistency in geographic
information,'' \emph{Information Sciences}, Vol. 176, No. 6, pp. 601--627, 2006.

\bibitem{Li2007} 
X.~Li, X.~Huang, J.~Dezert and F.~Smarandache, ``Enrichment of Qualitative Belief for Reasoning under Uncertainty,'' \emph{Proceedings of International Conference on Information Fusion, Fusion 2007}, Qu\' ebec, Canada, 9-12 July 2007.

\bibitem{Moratz}
R.~Moratz and M.~Ragni, ``Qualitative spatial reasoning about
relative point position ,''\emph{Journal of Visual Languages and
Computing}, In Press, Available online since January 25th, 2007.

\bibitem{Parsons1}
S.~Parsons and E.~Mamdani, ``Qualitative Dempster - Shafer theory'',
\emph{Proc. of the 3th EMACS Int. Workshop on Qualitative Reasoning and
Decision Technologies}, Barcelona, Spain,1993.

\bibitem{Parsons2}
S.~Parsons, ``Some qualitative approaches to applying Dempster-Shafer theory,''
 \emph{Information and Decision Technologies}, Vol. 19, pp.
321--337, 1994.

\bibitem{Parsons3}
S.~Parsons, ``A proof theoretic approach to qualitative probabilistic reasoning,''
 \emph{Int. J. of Approx. Reasoning,}, Vol. 19, No. 3-4, pp.
265--297, 1998.

\bibitem{Polya}
G.~Polya, ``Patterns of Plausible Inference'', \emph{Princeton University Press}, Princeton, NJ, 1954.

\bibitem{Ranganathan}
P.~Ranganathan, J.B.~Hayet, M.~Devy et al., ``Topological navigation and
qualitative localization for indoor environment using multi-sensory
perception,''\emph{Robotics and Autonomous Systems}, Vol. 49, No 1-2, pp. 25--42, 2004.

\bibitem{Shafer_1976}
G.~Shafer, ``A Mathematical Theory of Evidence", \emph{Princeton University Press}, Princeton, NJ, 1976.

\bibitem{SDBook2}
F.~Smarandache and  J.~Dezert (Editors), ``Applications and Advances of DSmT for Information Fusion (Collected works)", \emph{American Research Press}, Rehoboth, 2004.

{\small{http://www.gallup.unm.edu/{\verb+~+}smarandache/DSmT-book1.pdf}}.

\bibitem{SDBook}
F.~Smarandache and  J.~Dezert (Editors), ``Applications and Advances of DSmT for Information Fusion (Collected works)", Vol.2, \emph{American Research Press}, Rehoboth, 2006.

{\small{http://www.gallup.unm.edu/{\verb+~+}smarandache/DSmT-book2.pdf}}.

\bibitem{QBCR}
F.~Smarandache, J.~Dezert, ``Qualitative Belief Conditioning Rules (QBCR)", \emph{Proceedings of International Conference on Information Fusion, Fusion 2007}, Qu\' ebec, Canada, 9-12 July 2007.

\bibitem{Wagner}
T.~Wagner, U.~Visser and O.~Herzog, ``Egocentric qualitative spatial
knowledge representation for physical robots,''\emph{Robotics and
Autonomous Systems}, Vol. 49, No 1-2, pp. 25--42, 2004.

\bibitem{Wellman}
M.P.~Wellman, ``Some varieties of qualitative probability'', \emph{Proc.of the 5th Int. Conf. on Information Processing and the Management of Uncertainty (IPMU)}, Paris , July, 1994.

\bibitem{Wong}
S.K.M~Wong and P.~Lingras, ``Representation of qualitative user
preference by quantitative belief functions,''\emph{IEEE
Trans. on Knowlwdge and Data Engineering}, Vol.6, No.1, pp. 72--78, 1994.

\bibitem{Zadeh1}
L.~Zadeh, ''A Theory of Approximate Reasoning,''\emph{Machine
Intelligence},  Vol. 9, pp. 149--194, 1979.

\bibitem{Zadeh2}
L.~Zadeh, ``Fuzzy logic = Computing with words,''\emph{IEEE
Transactions on Fuzzy Systems}, Vol. 4, No. 2, pp. 103--111, 1996.


\end{thebibliography}
\end{document}